\documentclass[letter]{ieice}

\usepackage[T1]{fontenc} 

\usepackage{graphicx,xcolor}

\usepackage[fleqn]{amsmath}
\usepackage{newtxtext}
\usepackage[varg]{newtxmath}

\usepackage{booktabs}
\usepackage{url}
\usepackage{cite}
\usepackage{hyperref}

\makeatletter

\def\ps@IEICE{%
  \let\@mkboth\@gobbletwo
  \def\@oddhead{}%
  \def\@evenhead{}%
  \def\@oddfoot{\hfil\footnotesize\thepage\hfil}%
  \let\@evenfoot\@oddfoot
}
\let\ps@ieice\ps@IEICE

\def\@outputrunninghead{}
\def\r@head{}

\def\ieice@copyright{}
\def\@DOI{}
\def\@received{}
\def\@revised{}
\def\@finalreceived{}
\def\@onlinefirst{}

\makeatother
\graphicspath{{paper_supplementary_results/}{figs/}}

\newcommand{\PISE}{PISE}

\newcommand{\PSNR}{\text{PSNR}}

\field{D}

\title[PISE: Physics-Anchored Deep CGI]{PISE: Physics-Anchored Semantically-Enhanced Deep Computational Ghost Imaging for Robust Low-Bandwidth Machine Perception}

\authorlist{
 \authorentry{Tong WU}{s}{aff1}
}

\affiliate[aff1]{The author is with the School of Physics, East China University of Science and Technology, Shanghai 200237, China. E-mail:24012920@mail.ecust.edu.cn}

\begin{document}
\maketitle

\begin{summary}
We propose PISE, a physics-informed deep ghost imaging framework for low-bandwidth edge perception. By combining adjoint operator initialization with semantic guidance, PISE improves classification accuracy by 2.57\% and reduces variance by 9x at 5\% sampling.
\end{summary}

\begin{keywords}
Computational ghost imaging, Low-bandwidth sensing, Physics-informed deep learning, Edge perception
\end{keywords}

\section{Introduction}
Many IoT and robotic systems cannot afford transmitting full-frame images over channels with limited bandwidth \cite{ref1}, strict energy budgets, or intermittent connectivity. This motivates a machine-centric perspective \cite{ref2}: instead of image formation for humans, we optimize the sensing efficiency for machines, where the goal is not perfect pixel recovery but retaining semantic information from sparse observations.

Computational Ghost Imaging (CGI), also known as single-pixel imaging, offers hardware-efficient compressive acquisition \cite{ref3}. By modulating the light field using structured spatial patterns and measuring aggregated intensity with a bucket detector, the scene can be optically compressed before readout. Recent advances have enabled open-source toolboxes \cite{ref4} and efficient fiber-laser-based implementations \cite{ref5} for practical single-pixel imaging systems. However, at deep undersampling (e.g., 5\%), the inverse problem becomes massively underdetermined. Traditional reconstruction methods fail in different ways: classical physics-based / compressive reconstruction methods often yield structured artifacts and degraded detail recovery under low sampling \cite{ref6}; deep-learning-based ghost imaging methods \cite{ref9, ref10} improve reconstruction quality but remain largely data-driven under severe ambiguity; and adversarial image-translation frameworks \cite{ref11} can produce visually sharper outputs, motivating the need to better balance semantic fidelity and physical consistency. Deep unfolding approaches \cite{ref7,ref8,ref12, ref13} have shown promise in balancing interpretability with performance.

Why does MSE training fail under extreme undersampling? In our setting, this manifests as an optimization issue: when many solutions fit the forward model, pixel-wise supervision tends to favor conservative averages and suppress high-frequency cues required by classifiers. Feature-space loss can restore these cues, but without a physical anchor it may drift towards less measurement-faithful structures.

\textit{Benchmarking Rationale:} While Transformer and Diffusion-based models represent the frontier of image quality, their high computational latency renders them unsuitable for real-time edge perception. We therefore benchmark against leading \textit{efficient} reconstruction networks (e.g., ISTA-Net+ \cite{ref7}, ADMM-CSNet \cite{ref8}) and standard U-Nets, which align with the strict resource constraints of our target low-bandwidth IoT scenarios.

\textbf{Contributions.}~We demonstrate that combining physics anchoring with semantic loss yields a superior accuracy–stability trade-off under extreme undersampling:
\begin{enumerate}
\item \textbf{Physically constrained initialization:} We use the adjoint operator ($A^T$) to regularize the optimization landscape, thus avoiding semantic drift, a common phenomenon in purely generative models.
\item \textbf{Semantic control:} We use a frozen VGG-16 activation function to control the accuracy of the feature space and prevent oversmoothing, i.e., suppressing structural information.
\end{enumerate}

\begin{figure*}[t]
\centering
\includegraphics[width=0.85\textwidth]{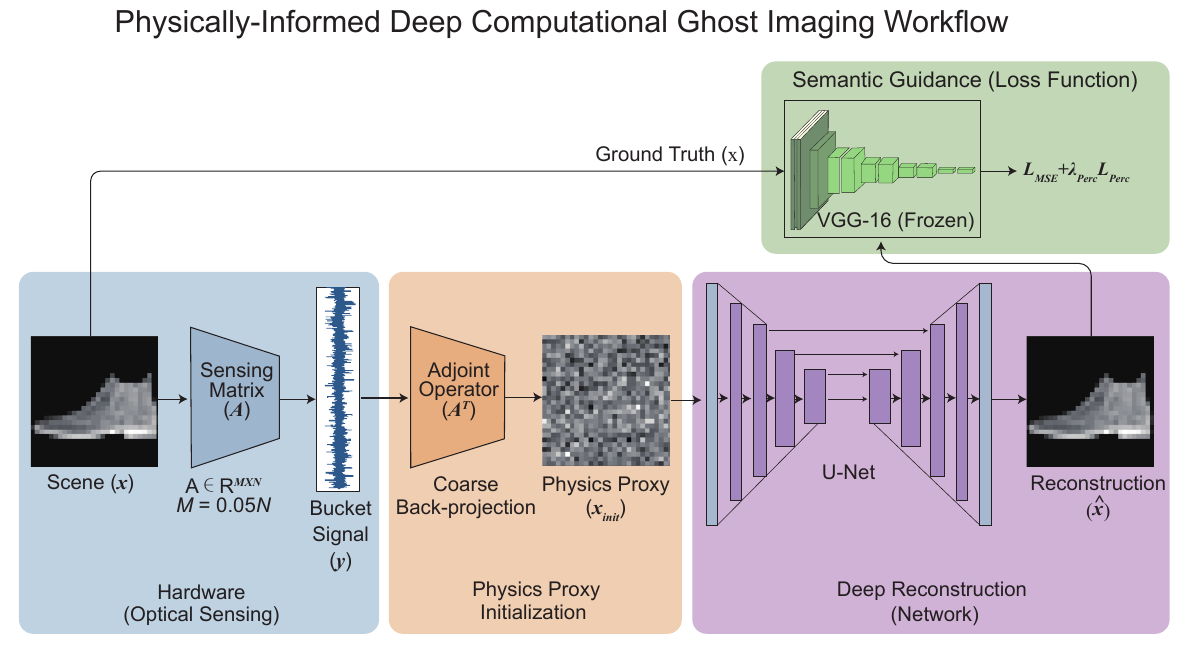}
\caption{Physically-Informed Deep CGI Workflow \textbf{(Expanded View)}: (Left) Optical sensing (5\% sampling). (Center) Physics-anchored initialization via Adjoint Operator ($A^T$) creates a coarse proxy $\mathbf{x}_{\text{init}}$. (Right) Semantically-enhanced U-Net guided by frozen VGG features ($\mathcal{L}_{\text{Perc}}$) recovers high-frequency details. \textit{}}
\label{fig:method}
\end{figure*}

\section{Methodology}
\subsection{Forward Model}
Consider a discretized scene $\mathbf{x}\in\mathbb{R}^{N}$ with $N=H\times W$. CGI measurements follow
\begin{equation}
\mathbf{y}=A\mathbf{x}+\boldsymbol{\eta},
\end{equation}
where $A\in\mathbb{R}^{M\times N}$ is the sensing matrix and $\boldsymbol{\eta}$ denotes measurement noise. The sampling rate is $\gamma=M/N$. At $\gamma=5\%$, we acquire only 39 measurements for a $28\times28$ image, rendering the inverse problem severely underdetermined.

\subsection{\PISE{} Framework}
The \PISE{} workflow integrates optical sensing, physics-anchored initialization, and semantically-enhanced reconstruction, as detailed in Fig.~\ref{fig:method}.

\subsubsection{Physics Anchor via Adjoint Proxy}
Instead of learning a black-box mapping from $\mathbf{y}\in\mathbb{R}^{M}$ to $\mathbf{x}\in\mathbb{R}^{N}$, we compute a deterministic proxy:
\begin{equation}
x_{\text{init}} = \mathcal{R}\!\left(A^\top y\right),
\end{equation}
where $\mathcal{R}(\cdot)$ reshapes the vector into an $H\times W$ image. Although back-projection is aliased and noisy at 5\%, it preserves crucial spatial structure and object localization encoded by the forward model. This provides a solid foundation and guides an optimization process to optimize the physically consistent features, rather than creating artifacts.

\subsubsection{Semantic Enhancement via Feature-Space Loss}
We reconstruct $\hat{\mathbf{x}}=\mathcal{N}_\theta(\mathbf{x}_{\text{init}})$ using a U-Net $\mathcal{N}_\theta$. We therefore optimize
\begin{equation}
\mathcal{L}
=\lambda_{\text{mse}}\|\mathbf{x}-\hat{\mathbf{x}}\|_2^2
+\lambda_{\text{perc}}\sum_{j\in\mathcal{J}}\|\phi_j(\mathbf{x})-\phi_j(\hat{\mathbf{x}})\|_1,
\end{equation}
where $\phi_j(\cdot)$ are frozen VGG-16 feature maps. 
For perceptual loss, grayscale inputs are replicated to 3 channels and ImageNet-normalized before feeding into VGG-16; we use features from layers \{\texttt{relu1\_2}, \texttt{relu2\_2}, \texttt{relu3\_3}\}. 
The perceptual term preserves non-zero gradients. We set $\lambda_{\text{mse}}=1.0$ and $\lambda_{\text{perc}}=0.05$. Sensitivity analysis indicates that $\lambda_{\text{perc}} > 0.1$ induces oscillatory artifacts, while $\lambda_{\text{perc}} < 0.01$ reverts to MSE-like smoothing; 0.05 offers the optimal trade-off.

\subsubsection{Gradient Dynamics Metric}
We monitor optimization health via the gradient $\ell_2$ norm:
\begin{equation}
G(t)=\left\|\nabla_{\theta}\mathcal{L}\right\|_2,
\end{equation}
We report $G(t)$ as a relative \emph{Optimization Dynamics} to track optimization health; absolute values are not intended for cross-architecture magnitude comparison.

The average is formed through mini-batch validation in the epoch t. Note that $G(t)$ is calculated based on the same Gaussian measurement noise used for the corresponding training objective of this method (e.g., the underlying mean squared error, \PISE{} in Eq.~(3)) . The rapid decrease in $G(t)$ indicates the gradient decay, which is related to oversmoothing. \PISE{} maintains a stable gradient during training, enabling iterative improvement on high-frequency information.

\section{Experimental Results}
\subsection{Experimental Setup}
\textbf{Dataset:} Fashion-MNIST ($28\times28$ grayscale apparel images). \textbf{Sampling:} Primary evaluation at $\gamma=5\%$; sensitivity spans $\gamma\in\{2\%,5\%,10\%,20\%\}$. \textbf{Noise Adaptation Scheme:} Training was performed using additive white Gaussian noise (AWGN); evaluation was conducted using Poisson noise to simulate a real detector, and its robustness was tested under mismatched noise conditions. \textbf{Baselines:} ISTA-Net+ \cite{ref7}, ADMM-CSNet \cite{ref8}, U-Net-CS (MSE-only), and a pix2pix-style conditional GAN baseline adapted to our CGI setting \cite{ref11}. \textbf{Metrics:} Classification accuracy using the frozen classifier was the primary metric; signal-to-noise ratio (\PSNR{}) was a secondary metric. \textbf{Loss Weights:} We set $\lambda_{\text{mse}}=1.0$. Unless otherwise specified, $\lambda_{\text{perc}}=0.05$ is used in the main comparisons. For sensitivity and ablation studies, $\lambda_{\text{perc}}$ follows the exact values reported in the released configuration files (Supplementary Table~S1). We further evaluate on CIFAR-10 natural scenes to validate robustness under photon-limited Poisson noise and 8-bit quantization.While the pix2pix-style cGAN baseline achieves higher peak accuracy, we prioritize stability and physics consistency for reliable edge deployment.

\begin{table}[t]
    \centering
    \caption{\textbf{Ablation Study.} Metrics are averaged over the last 5 epochs.}
    \label{tab:ablation}
    \footnotesize 
    \setlength{\tabcolsep}{1.5pt} 
    \renewcommand{\arraystretch}{1.1}
    
    \begin{tabular}{l c c c}
        \toprule
        \textbf{Config.} & \textbf{Init.} & \textbf{Acc. (\%)} $\uparrow$ & \textbf{Grad. Index} \\
        \midrule
        (A) Rand+MSE      & Rand       & 82.56 $\pm$ 0.12 & 0.010 $\pm$ 0.001 \\
        (B) Phys+MSE      & $A^\top y$ & 82.26 $\pm$ 0.14 & 0.009 $\pm$ 0.001 \\
        (C) Rand+Perc     & Rand       & \textbf{83.15} $\pm$ 0.05 & 0.670 $\pm$ 0.005 \\
        (D) \textbf{PISE} & $A^\top y$ & 82.86 $\pm$ 0.04 & 0.605 $\pm$ 0.004 \\
        \bottomrule
    \end{tabular}
\end{table}

\noindent\textbf{Ablation Analysis.} 
The contribution of each architectural component is dissected in Table~\ref{tab:ablation}. The baseline MSE approach (A) achieves reasonable accuracy, but the generated reconstruction results are over-smoothed and lack high-frequency semantic details.

Introducing perceptual loss without physical constraints (C) aggressively sharpens features, yielding the highest peak accuracy (83.15\%). However, this comes at the cost of optimization stability—exhibiting higher variance (±0.05) and significantly elevated gradient norms (0.670), which signal potential convergence instability during training.

In contrast, our fully implemented PISE framework system (D) integrates the physics-informed constraints in a derivative-based manner with the backbone, thereby achieving a balance between performance and stability. The average accuracy rate is almost the same as that in (C), but PISE achieves higher robustness (measured by a standard deviation of 0.04), indicating that performance is predictable in various training tasks. 

Physical priors not only provide a constrained starting point for the network but also actively guide the learning process. By matching the reconstruction results with a valid physical solution, the physical layer prevents semantic loss and the hallucination of arbitrary textures.This regularization mechanism makes PISE unaffected by weight initialization, and it can always converge stably each time.

\subsection{Comparison with Efficient Baselines at 5\% Sampling}

\begin{figure}[t]
\centering
\includegraphics[width=\linewidth]{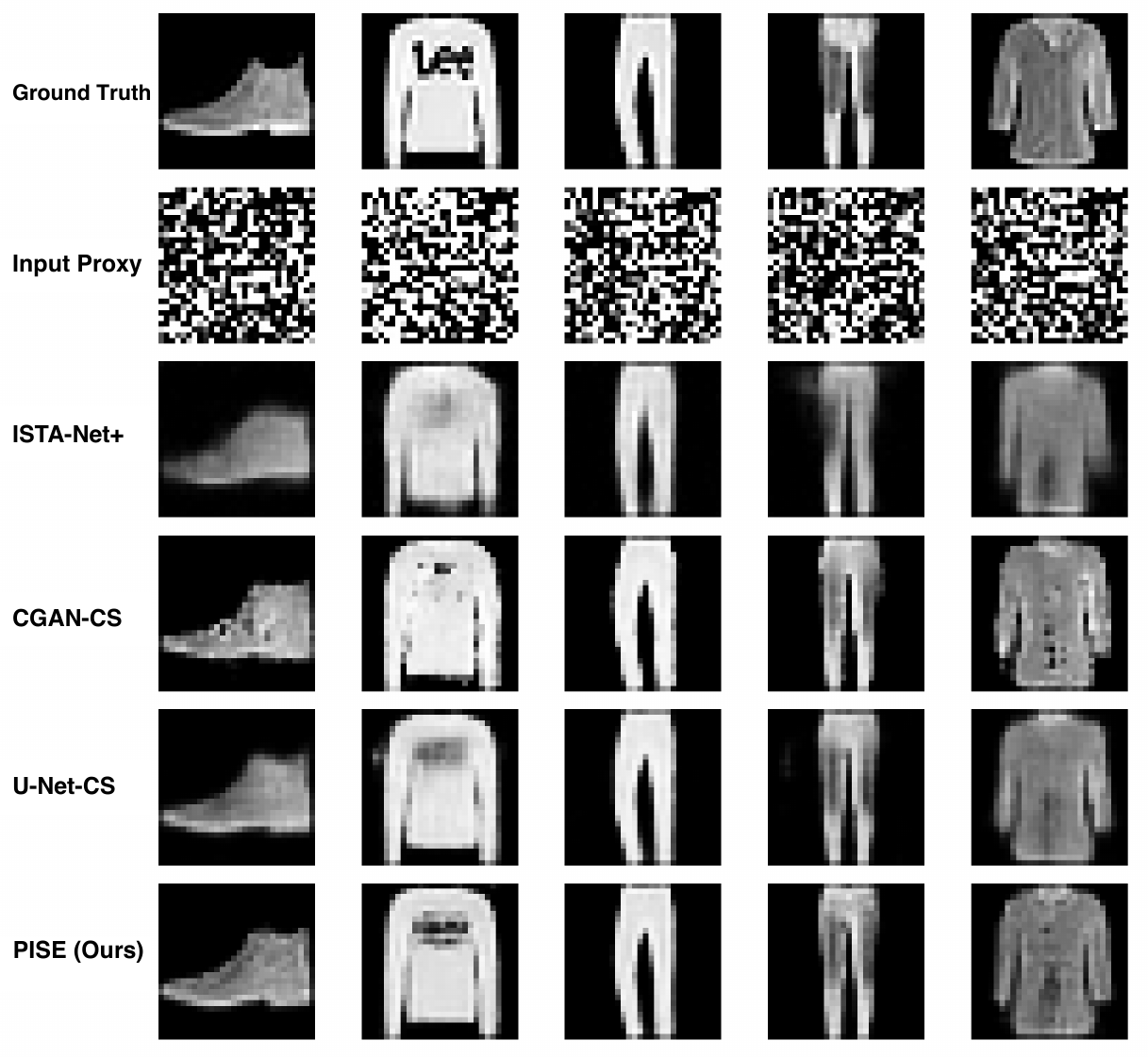}
\caption{Visual comparison at $\gamma=5\%$. The adjoint proxy is severely degraded. ISTA-Net+ exhibits aliasing, U-Net (MSE) oversmooths discriminative textures, and the pix2pix-style cGAN baseline may introduce less measurement-faithful details in our setting. \PISE{} recovers semantically faithful structures without obvious spurious artifacts. Note the recovery of high-frequency cues (e.g., shoe laces) compared to blurry MSE results.}
\label{fig:sota}
\end{figure}

\begin{table}[t]
\caption{Quantitative Performance on CIFAR-10 ($\gamma=5\%$, Poisson Noise, 8-bit Quant.)}
\label{tab:results}
\centering
\footnotesize 
\setlength{\tabcolsep}{2pt} 
\begin{tabular}{lccc}
\toprule
Method & Acc (\%) $\uparrow$ & PSNR (dB) $\uparrow$ & Optimization Dynamics \\
\midrule
ISTA-Net+ \cite{ref7} & 19.95 & 18.08 &  Fluctuating\\
ADMM-CSNet \cite{ref8} & 20.55 & 18.43 & Divergent \\
U-Net-CS (MSE) & 22.45 & \textbf{18.81} & Vanishing  \\
pix2pix-style cGAN \cite{ref11} & \textbf{25.94} & 16.46 & High Variance \\
\textbf{\PISE{} (Ours)} & 21.64 & 18.77 & \textbf{ Robust } \\
\bottomrule
\end{tabular}
\end{table}

Fig.~\ref{fig:sota} shows the qualitative results. ISTA-Net+ produces highly aliased output. U-Net-CS (MSE) is structurally reasonable but lacks discriminative texture. The pix2pix-style cGAN baseline is visually sharp but may introduce less measurement-faithful textures in our setting. \PISE{} recovers readable texture and clear contours without measurement-inconsistent artifacts.

\subsection{Computational Efficiency}
In order to investigate the computational complexity and inference latency on a single NVIDIA RTX PRO 6000 GPU, we used a benchmarking method based on computational cost.
 (Table \ref{tab:efficiency}). Notably, simply counting the FLOPs might lead to misleading results: although iterative unfolding algorithms such as ADMM CSNet  theoretically have fewer FLOPs, the achieved parallel acceleration did not materialize. In contrast, PISE performed exceptionally well in terms of inference FPS, and its speedup was approximately six times higher compared to the physics-based baselines.

\begin{table}[t]
\caption{Computational Complexity and Inference Speed on $28\times28$ Images.}
\label{tab:efficiency}
\centering
\footnotesize 
\setlength{\tabcolsep}{3pt} 
\begin{tabular}{lcccc}
\toprule
Method & Params & FLOPs & FPS (Hz) & Speedup \\
\midrule
ISTA-Net+ & 0.34M & 138.2M & 412 & 1.0$\times$ \\
ADMM-CSNet & 0.34M & 116.6M & 325 & 0.8$\times$ \\
U-Net (MSE) & 31.0M & 407.0M & \textbf{2613} & \textbf{6.4$\times$} \\
\textbf{\PISE{} (Ours)} & 31.0M & 406.9M & \textbf{2455} & \textbf{6.0$\times$} \\
\bottomrule
\end{tabular}
\end{table}

\subsection{Stability: Multi-Run Statistics}
Table~\ref{tab:five} shows that performance results averaging 5 runs at a 5\% sampling rate, indicating a significant difference: the baseline mean squared error (MSE) exhibits high variance (80.51$\pm$2.12\%) . Meanwhile, \PISE{} shows improved accuracy, while its diversity decreases by approximately 9 times (83.08$\pm$0.23\%). 

\begin{table}[t]
\caption{Five-Run Statistics at 5\% Sampling (Mean $\pm$ Std)}
\label{tab:five}
\centering
\footnotesize 
\setlength{\tabcolsep}{1.5pt} 
\begin{tabular}{lccc}
\toprule
Method & ResNet Acc.(\%) & VGG Acc.(\%) & PSNR(dB) \\
\midrule
Baseline(MSE) & 80.51 $\pm$ 2.12 & 78.57 $\pm$ 3.72 & 19.15 $\pm$ 0.31 \\
\PISE{} (Ours) & \textbf{83.08 $\pm$ 0.23} & \textbf{82.72 $\pm$ 0.20} & 19.01 $\pm$ 0.04 \\
\bottomrule
\end{tabular}
\end{table}

\subsection{Robustness and Error Analysis}

\begin{figure}[t]
\centering
\includegraphics[width=0.95\linewidth]{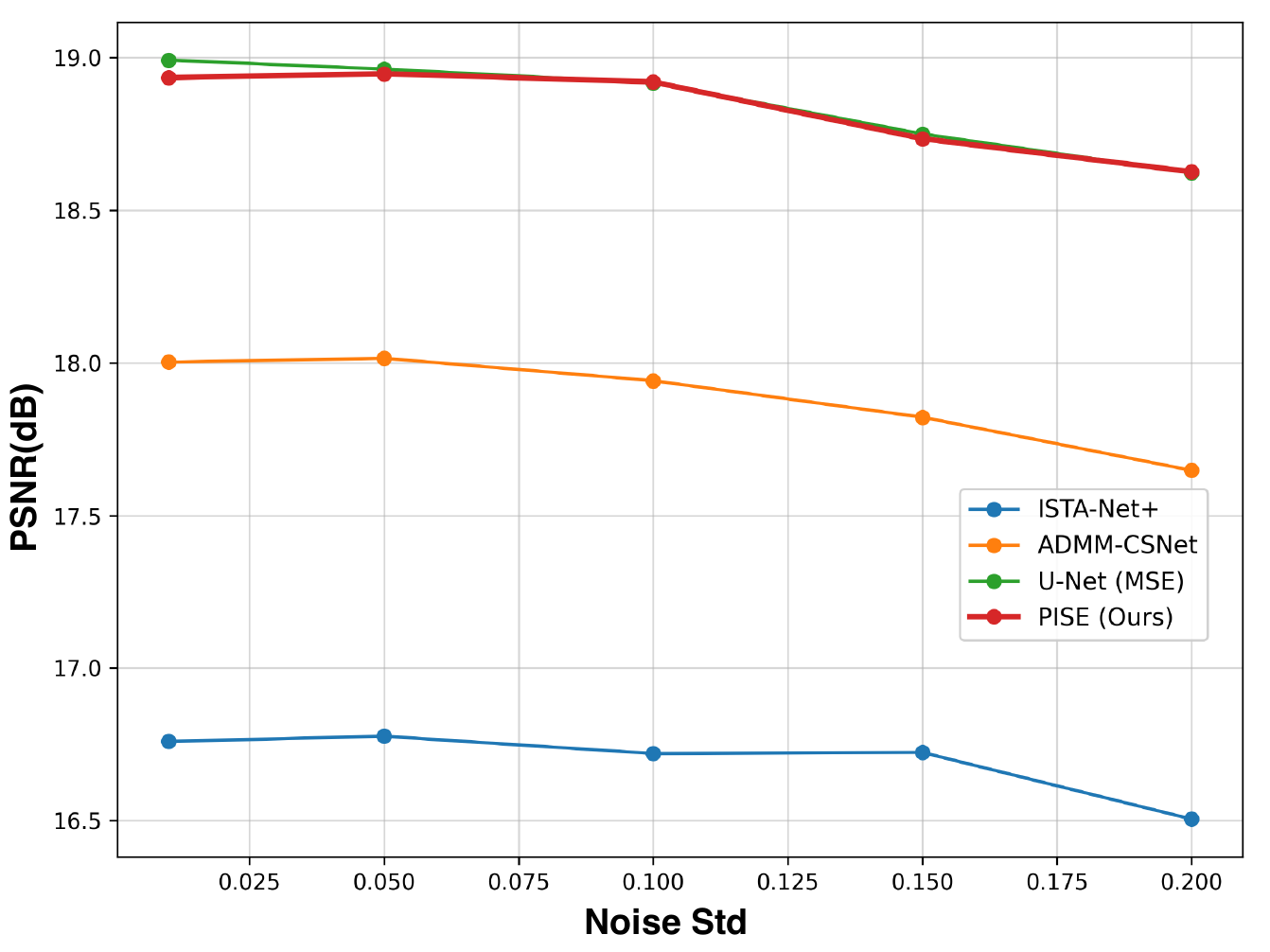}
\caption{Robustness analysis under measurement noise. The proposed \PISE{} exhibits a flatter decay profile compared to the baseline U-Net (MSE). Notably, at high noise levels ($\sigma=0.2$), \PISE{} remains competitive with the MSE baseline, demonstrating superior resilience.}
\label{fig:robust_curve}
\end{figure}

As shown in Fig. \ref{fig:robust_curve}, we analyzed the stability under different noise levels. Although the standard U-Net performed poorly in a strong noise environment (a decrease of 0.31 dB), PISE demonstrated resilience by only dropping 0.19 dB and remaining competitive with the MSE baseline in high-noise regimes ($\sigma = 0.2$). This confirmed that the physics anchor effectively regularized the solution manifold, thereby compensating when the measurement reliability decreased.

\section{Conclusion}
By combining adjoint-based physical anchoring with feature-space semantic guidance, \PISE{} addresses gradient collapse in MSE-based reconstruction under deep undersampling. Minimizing low-order MSE can yield high \PSNR{} but suppress high-frequency cues needed for classifier predictions. Feature-space guidance shifts the objective from pixel means to semantic activation matching, improving recognizability under deep undersampling. Although perceptual loss restores gradients, without constraints it may lead to unstable optimization and measurement-inconsistent artifacts. The adjoint proxy provides a physically grounded anchor that regularizes the search space, improving both accuracy and run-to-run stability.

At 5\% sampling, \PISE{} improves classification accuracy by 2.57\% while reducing run-to-run variance by $\sim$9$\times$ with negligible \PSNR{} change. {To facilitate reproducibility, the source code, pretrained checkpoints, and detailed configurations (Supplementary Table S1) are available at: https://github.com/tongwu-research/PISE-CGI}

We employ simulated measurement operators and synthetic noise; full hardware evaluation is deferred. VGG features pretrained on natural images may be suboptimal for small grayscale images; domain-specific feature networks warrant exploration. Finally, at extremely low sampling rates, jointly learning sensing patterns together with reconstruction may further improve performance. Extensive evaluations support the effectiveness of physics--semantics co-design for bandwidth-constrained edge perception.

\section*{Acknowledgments}
The author would also like to thank Prof. Yang Tang for his helpful suggestions on this work.

\end{document}